\documentclass[3p, preprint, 12pt]{elsarticle}

\usepackage{graphicx}
\usepackage{amssymb}

\usepackage{lineno}
\usepackage{threeparttable}
\usepackage{amsmath}
\usepackage{mathtools}
\usepackage{bm}
\usepackage{subcaption}
\usepackage{booktabs}
\usepackage{hyperref}
\usepackage{multirow}
\usepackage{xcolor}
\usepackage[ruled,vlined]{algorithm2e}
\usepackage{enumitem}
\usepackage{makecell}
\usepackage{xcolor}

\begin{document}

\begin{frontmatter}


%
\title{Exploring Large Language Models for \\Human Mobility Prediction under Public Events}

\author[a,b]{Yuebing Liang} 
\author[c]{Yichao Liu}
\author[a]{Xiaohan Wang}
\author[a,d,e]{Zhan Zhao\corref{cor1}}

\address[a]{Department of Urban Planning and Design, The University of Hong Kong, Hong Kong SAR, China}
\address[b]{Senseable City Lab, Massachusetts Institute of Technology, Cambridge, MA 02139, United States}
\address[c]{School of Architecture, Tsinghua University, Beijing, China}
\address[d]{Urban Systems Institute, The University of Hong Kong, Hong Kong SAR, China}
\address[e]{Musketeers Foundation Institute of Data Science, The University of Hong Kong, Hong Kong SAR, China}

\cortext[cor1]{Corresponding author (zhanzhao@hku.hk)}

\begin{abstract}
Public events, such as concerts and sports games, can be major attractors for large crowds, leading to irregular surges in travel demand. Accurate human mobility prediction for public events is thus crucial for event planning as well as traffic or crowd management.
While rich textual descriptions about public events are commonly available from online sources, it is challenging to encode such information in statistical or machine learning models. Existing methods are generally limited in incorporating textual information, handling data sparsity, or providing rationales for their predictions.
To address these challenges, we introduce a framework for human mobility prediction under public events (LLM-MPE) based on Large Language Models (LLMs), leveraging their unprecedented ability to process textual data, learn from minimal examples, and generate human-readable explanations. Specifically, LLM-MPE first transforms raw, unstructured event descriptions from online sources into a standardized format, and then segments historical mobility data into regular and event-related components. A prompting strategy is designed to direct LLMs in making and rationalizing demand predictions considering historical mobility and event features. A case study is conducted for Barclays Center in New York City, based on publicly available event information and taxi trip data. Results show that LLM-MPE surpasses traditional models, particularly on event days, with textual data significantly enhancing its accuracy. Furthermore, LLM-MPE offers interpretable insights into its predictions. Despite the great potential of LLMs, we also identify key challenges including misinformation and high costs that remain barriers to their broader adoption in large-scale human mobility analysis.

\end{abstract}

\begin{keyword}
Public Events \sep Large Language Models \sep Human Mobility Prediction \sep Travel Demand Modeling \sep Text Data Mining


\end{keyword}

\end{frontmatter}


\section{Introduction} \label{sec:intro}
Public events, such as concerts, sports games and conventions, are scheduled events at pre-determined venues that may be attended by the general public. They play an important role in fostering cultural exchange, promoting social inclusion, boosting urban vitality, and enriching local economy. As these events often attract a large number of visitors to a specific location at a specific time, they are also a key contributor to irregular surges in travel demand, leading to potential transportation and urban challenges like traffic congestion \citep{chang2013multinomial}, overcrowded public transportation \citep{rodrigues2016bayesian}, and safety risks such as stampedes or road accidents \citep{li2017forecasting}. To maintain safety in event venues and ensure the smooth functioning of transportation systems, it is crucial to accurately predict travel demand patterns for these events in advance. Based upon this, city administrators and transportation system operators can proactively implement strategies including modifying taxi dispatches, increasing public transit frequencies, introducing additional bus feeders, and broadcasting early warnings \citep{li2017forecasting}.

In the past decades, numerous methods have been proposed for travel demand prediction \citep{zhao2022coupling, liang2022joint}, but they usually overlook the irregular mobility patterns resulting from public events. Modeling human mobility under public events presents distinct challenges due to the complex impact of such events on travel patterns. 
For large-scale events like the Olympics or World Cups, event organizers and city planners tend to rely on extensive manual forecasting, planning years in advance \citep{pereira2015using}. Although these approaches are thorough, they are time-consuming and not scalable to the variety of smaller events frequently taking place in cities. Although some recent studies attempted to infer irregularities from historical mobility data \citep{chen2019subway, zhao2022coupling, guo2023new}, they struggled to identify the precise timings of events and their unique attributes. 

The widespread adoption of the internet and social media platforms has provided a fresh avenue for gathering event details. Event organizers commonly advertise their events on ticketing websites and social media, highlighting essential details such as date, time, location, and textual descriptions. Prior studies have demonstrated the potential of such online information in enhancing mobility prediction accuracy under special events \citep{ni2016forecasting}. However, most of them only integrated numerical or categorical data, including event categories \citep{pereira2015using}, timing \citep{tu2023multi}, and the number of related social media posts \citep{ni2016forecasting, xue2022forecasting}. The rich textual information of event descriptions, which can provide valuable insights about event themes, participants and background details, is usually ignored. Only a few recent studies have attempted to leverage textual data using techniques such as topic modeling \citep{rodrigues2016bayesian} and word embeddings \citep{rodrigues2019combining}, but several key challenges remain.

The first challenge involves effectively utilizing the rich textual data from online event descriptions that often vary in structure and depth. Some events come with minimal text descriptions, necessitating deeper semantic understanding by models. For example, an event description like ``WALK THE MOON" may not be readily interpreted by traditional language models as a concert featuring the American rock band named \textit{Walk the Moon}. Conversely, for events described in extensive details, models need to sift through the description to extract pertinent information. For example, if an event description delves into a performer's life story, it would be challenging for language models to distinguish between extraneous details and relevant information for mobility prediction.

The second challenge lies in data sparsity resulting from the infrequent and irregular occurrence of events. For example, a prior study highlighted that, in an 8-month period from 2017-10-27 to 2018-05-31, only 11 events took place in the studied location (i.e., Beijing Workers' Stadium) \citep{tu2023multi}. Data sparsity complicates the analysis of human mobility trends, especially considering the distinct impact each event may have. For instance, the travel demand generated by two different concerts could vary greatly, largely influenced by the performer's popularity level. It is thus particularly challenging to estimate the impact of future events with minimal historical data. 

The third challenge pertains to deciphering the underlying logic of model predictions. While machine learning and deep learning models have showcased impressive predictive prowess, their intrinsic ``black box" nature has drawn criticism due to their lack of transparency and interpretability. For event planning and traffic management tasks, having interpretable models is crucial not only for confirming the reliability of their outcomes but also for providing useful insights for transportation system operators in their decision-making processes. Despite recent efforts employing explainable AI techniques to discern the influence of spatiotemporal features on demand prediction \citep{liang2023cross}, the resultant interpretations remain somewhat restricted and are not easily understood by humans.

Recently, large language models (LLMs), such as GPT-4 and Llama, have been recognized as powerful tools for a variety of natural language processing tasks. These models contain tens to hundreds of billions of parameters and are pre-trained over large-scale corpora from the internet. Their vast scale, combined with the extensive training data, equips them with emergent abilities not seen in smaller models \citep{zhao2023survey}. 
First, due to their extensive training on wide-ranging internet content, LLMs possess an encyclopedic knowledge base \citep{yu2023temporal}, making them adept at handling varied online event descriptions, regardless of their format or length. Second, LLMs are equipped with a ``few-shot learning" ability, which means they can undertake new tasks with minimal examples or simple instructions \citep{brown2020language}. 
Third, LLMs have the inherent ability to generate human-understandable natural language texts, and with proper prompting methods, they can produce justifications for their predictions \citep{yu2023temporal}. Therefore, LLMs stand as a promising solution to all the aforementioned challenges. 

In this research, we introduce a framework for human mobility prediction under public events based on LLMs, termed LLM-MPE. The LLM-MPE approach is divided into two primary phases: (1) feature development and (2) mobility prediction. In the feature development phase, we utilize two types of data: past event descriptions and human mobility flows. To ensure the model effectively understands the influence of public events on travel demand, we decompose past human mobility flows into regular patterns (largely resulting from daily commutes) and irregular deviations (largely resulting from public events). Moreover, we employ LLMs to standardize event description data of varied structures and lengths into a uniform and succinct format. For the mobility prediction phase, we create efficient prompts to guide LLMs in inferring the likely travel demand for upcoming public events, while also elucidating the rationale behind its predictions. The model is demonstrated through a case study on Barclays Center in New York City (NYC) with multi-source data. 
To encapsulate, our study makes contributions in the following aspects:

\begin{itemize}[noitemsep]
    \item A LLM-based human mobility prediction framework is developed for public events, seamlessly integrating historical mobility patterns with textual event descriptions. To the best of our knowledge, this study represents the first attempt to harness LLMs for human mobility prediction under special events.  
    \item Through a case study of Barclays Center in NYC with publicly available event data and taxi trip records, we empirically show the value of incorporating textual descriptions of events, and that LLM-MPE is able to outperform existing machine learning and deep learning models by a large margin. 
    \item We employ a chain-of-thought prompting strategy to make LLM-MPE detail its decision-making progressively. This offers clarity on the model's predictive reasoning, enhancing its trustworthiness and providing valuable insights for managing public events.
    \item To guide future research in improving LLM applications for human mobility prediction, we offer an in-depth analysis regarding the current limitations of LLMs, including generating misleading information, dependence on outdated knowledge, high costs of operation, constraints in prediction efficiency, and limited capacity to comprehend spatial relationships.
\end{itemize}






\section{Literature Review} \label{sec:literature}

\subsection{Human Mobility Forecasting for Public Events}
Transport planning traditionally focuses on only large-scale events like the Olympics and World Cups, requiring extensive manual work and collaboration between public and private sectors \citep{coutroubas2003public, vougioukas2008transport}. In modeling demand for these events, the classical four-step modeling approach is commonly used, relying heavily on travel survey data. For instance, \cite{kuppam2013special} conducted a survey at event venues in Phoenix, Arizona, gathering data from nearly 6,000 individuals for 20 events. The data was then used to predict the impact of special events on various travel indicators, including travel demand by mode, trip origin-destination patterns, and more. Similarly, \cite{chang2013multinomial} surveyed around 1,000 concertgoers in Taipei to develop a model explaining their travel mode and arrival time choices. In Istanbul, Turkey, \cite{e2021planned} surveyed soccer game attendees and adapted the traditional four-step model for a custom demand model under special events. While these methods provide useful tools for event-specific transport planning, the need to collect travel survey data can be time-consuming and labor-intensive, limiting the scalability of these methods to the large number of smaller events that frequently take place in cities.

Some studies have focused on improving special event human mobility prediction by identifying potential anomalies from historical mobility patterns. To forecast subway passenger flow during special events,  \cite{li2017forecasting} proposed a multiscale radial basis function network, using demand fluctuations from other subway stations as indicators of potential passenger surges at the target station. \cite{chen2019subway} proposed a hybrid approach based on smart card data, utilizing ARIMA and GARCH models to estimate the average and variability of passenger flows respectively. A naive Bayes-based transition model was established by \cite{zhao2022naive}, which first assessed the likelihood of special events and then applied different sub-models for demand predictions under regular and special event scenarios. \cite{guo2023new} introduced a Markov chain method for individual mobility forecasting that used collective mobility data to identify potential large-scale crowding events. However, a significant limitation of these approaches is their reliance solely on historical mobility data. Although these methods can indicate potential anomalies in demand, they lack the ability to accurately identify the occurrence of events or consider their specific characteristics.

Recent research efforts have highlighted that online platforms, such as announcement websites and social media, can be useful resources for identifying events and gathering comprehensive information about them, including dates, times, locations, and detailed descriptions \citep{xu2018sensing}. This insight has steered research toward integrating online data into human mobility prediction models. \cite{pereira2015using} built a deep neural network for forecasting public transport usage based on website-derived event categories, which, however, are often unavailable as noted by \cite{rodrigues2019combining}, making this method hard to generalize.
\cite{tu2023multi} gathered event timings from ticketing websites and used them with historical demand data in a deep learning model. Meanwhile, other studies have turned to social media data. \cite{ni2016forecasting} used the social media post volume and user count as event indicators, and designed a hybrid model to integrate these features with historical mobility data. Similarly, \cite{xue2022forecasting} incorporated social media post trends and nearby station flow disturbances using a hybrid deep neural network approach. These methods generally transform event information into numerical or categorical features, overlooking that a substantial amount of event information, such as descriptions from announcement websites and social media posts, exists in the unstructured textual form \citep{rodrigues2019combining}. Nonetheless, due to the disparity of data formats, it is difficult to directly integrate such textual data with existing statistical and machine learning models.

Only limited research attempts have incorporated textual data into demand prediction for special events. \cite{rodrigues2016bayesian} applied a Latent Dirichlet Allocation (LDA) model to transform website event descriptions into topic distribution vectors, which were then merged with transit smart card data in a Bayesian model with Gaussian processes. \cite{rodrigues2019combining} developed a deep learning approach, using pre-trained word embeddings, convolutional layers, and attention mechanisms to combine text information with demand data. While these methods are good at grasping individual word meanings, they might struggle to fully understand the semantics of complete paragraphs, especially when event descriptions are messy, incomplete, or irrelevant. For instance, they might not recognize ``Brooklyn Nets vs. Atlanta Hawks" as an NBA basketball game due to the absence of explicit keywords. Therefore, there is a need for a novel approach that can better interpret the semantics of textual data for enhanced demand prediction during special events.

\subsection{Large Language Models}

Large Language Models like GPT-4 \citep{OpenAI2023GPT4TR} and Llama \citep{touvron2023llama} are advanced Transformer-based models consisting of tens to hundreds of billions of parameters and trained on extensive text corpora \citep{zhao2023survey}. The substantial size and diverse training data of these models endow them with unique capabilities not found in smaller models. These include in-context few-shot learning, which allows them to understand and perform tasks with minimal examples \citep{brown2020language}, and step-by-step reasoning, enabling them to progressively solve complex problems \citep{wei2022chain}. 

The great success of LLMs has spurred significant interest in their application across urban and transportation studies. Previous research has primarily focused on assessing LLMs’ capacity in language-centric tasks, such as comprehending urban concepts \citep{fu2023towards}, responding to geospatial inquiries  \citep{mooney2023towards}, recognizing location descriptions \citep{mai2023opportunities}, and extracting information from accident narratives \citep{mumtarin2023large}. Simultaneously, several studies have harnessed LLMs to facilitate human-machine interactions. 
For instance, \cite{zhang2023geogpt} used LLMs to call established GIS tools to aid non-professional users in handling geospatial tasks. In a similar vein, \cite{zhang2024semantic} incorporated LLMs into a deep learning-based traffic data imputation system, facilitating user queries in simple language. 
While most existing efforts employ LLMs to enhance language tasks or natural language interactions for non-experts, \cite{wang2023would} used LLMs as human mobility predictors, offering precise and interpretable forecasts about an individual’s next location. However, they concentrated on routine and repetitive movement patterns. The potential of LLMs to tackle more complex human mobility prediction tasks, especially in scenarios involving public events, has yet to be explored.

\section{Methodology} \label{sec:method}

\subsection{Definitions and Problem Statement}

\textit{Definition 1 (Travel Demand in an Event Venue):}  For a given event venue $v$ at a particular time step $t$, we measure its travel demand as the number of outflow (departure) and inflow (arrival) trips in the vicinity, denoted by a two-dimensional vector $Y_{v,t} \in \mathbb{R}^2$. 

\textit{Definition 2 (Public Event):} A public event is represented as $e = (t_e, v_e, h_e)$, with $t_e$ specifying the event time, $v_e$ indicating the venue, and $h_e$ encompassing the event's features in unstructured text format. In our context, these features include event titles (and descriptions when available) obtained from online sources. For instance, a public event might be (2015-05-01 7:30 PM-10:30 PM, Barclays Center, ``Brooklyn Nets vs. Dallas Mavericks''), denoting an NBA basketball match between Brooklyn Nets and Dallas Mavericks at the Barclays Center on 2015-05-01 from 7:30 PM to 10:30 PM. The event features are further mapped to time intervals. Specifically, given an event venue $v$ and time interval $t$, the related event features are defined as $E_{v,t} = \{e_1, e_2, ..., e_c\}$, indicating that there are $c$ events at venue $v$ and time $t$, each characterized by their respective features.

\textit{Problem (Human Mobility Prediction in an Event Venue):} For an event venue $v$, our goal is to predict the upcoming travel demand denoted as $Y_{v,t+1}$, based on historical mobility data from the last $T$ time periods, $Y_{v,t-T:t}$, combined with the features of past events, $E_{v,t-T:t}$, and the features of the next scheduled event(s), $E_{v,t+1}$. This can be formulated as: 
\begin{equation}
    Y_{v,t+1} = F(Y_{v,t-T:t}, E_{v,t-T:t}, E_{v,t+1}),
\end{equation}
where $F(\star)$ represents the predictive model to be developed. Following \cite{rodrigues2019combining}, we will focus on predicting daily travel demand to ensure there is ample time for organizing transportation before events, as suggested by \cite{pereira2015using}. However, the methodology can be easily adapted for different time interval settings. 

\subsection{Overview}

\begin{figure}[ht!]
  \centering
  \includegraphics[width=\linewidth]{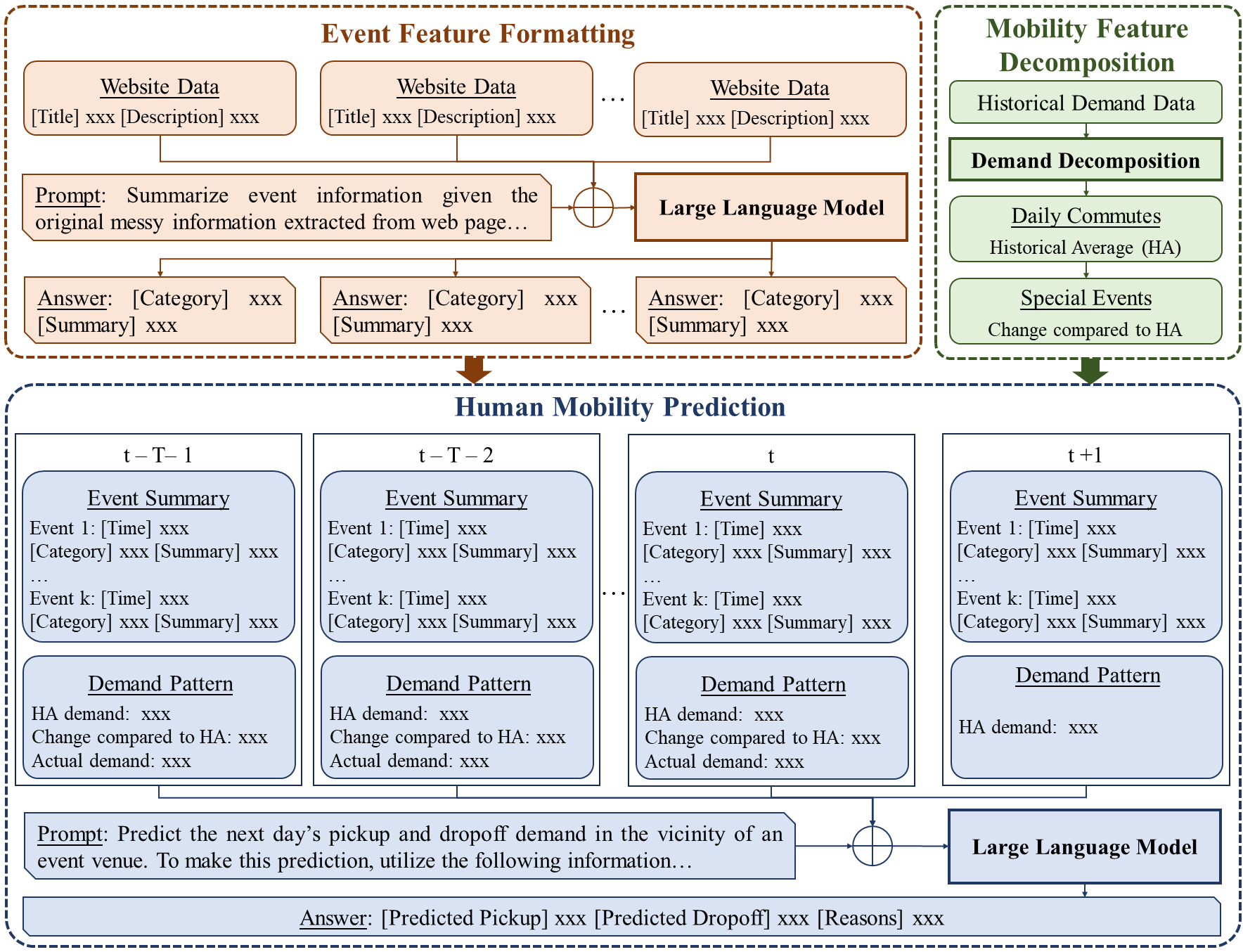}
  \caption{The framework of LLM-MPE} 
\label{fig:framework}
\end{figure}

LLM-MPE consists of three primary stages, as depicted in Figure~\ref{fig:framework}: (1) \textit{event feature formatting}, where LLMs transform unstructured event data from online sources into a standardized and succinct format; (2) \textit{mobility feature decomposition}, which separates regular mobility patterns from irregular deviations due to public events; and (3) \textit{human mobility prediction}, where LLMs produce forecasts and explanations for travel demand predictions using the processed event and mobility features. We provide a detailed explanation of each stage in the following sections.

\subsection{Event Feature Formatting}\label{method:event}
As mentioned in Section~\ref{sec:intro}, raw event data from online sources often presents inconsistencies in the format and length, making it a challenge for conventional language models to comprehend their semantics. This complexity arises especially when the event data is either cluttered with superfluous details or lacks essential information. LLMs excel in distilling extensive event details into concise and pertinent summaries due to their ability to generate natural language. Moreover, they are capable of identifying specific terms like the performer's name or show's title, because LLMs are pre-trained based on vast web data (e.g., Wikipedia) \citep{zhao2023survey}, enriching them with expansive real-world knowledge. This allows LLMs to deduce event details with scant data.

\begin{figure}[ht!]
  \centering
  \includegraphics[width=\linewidth]{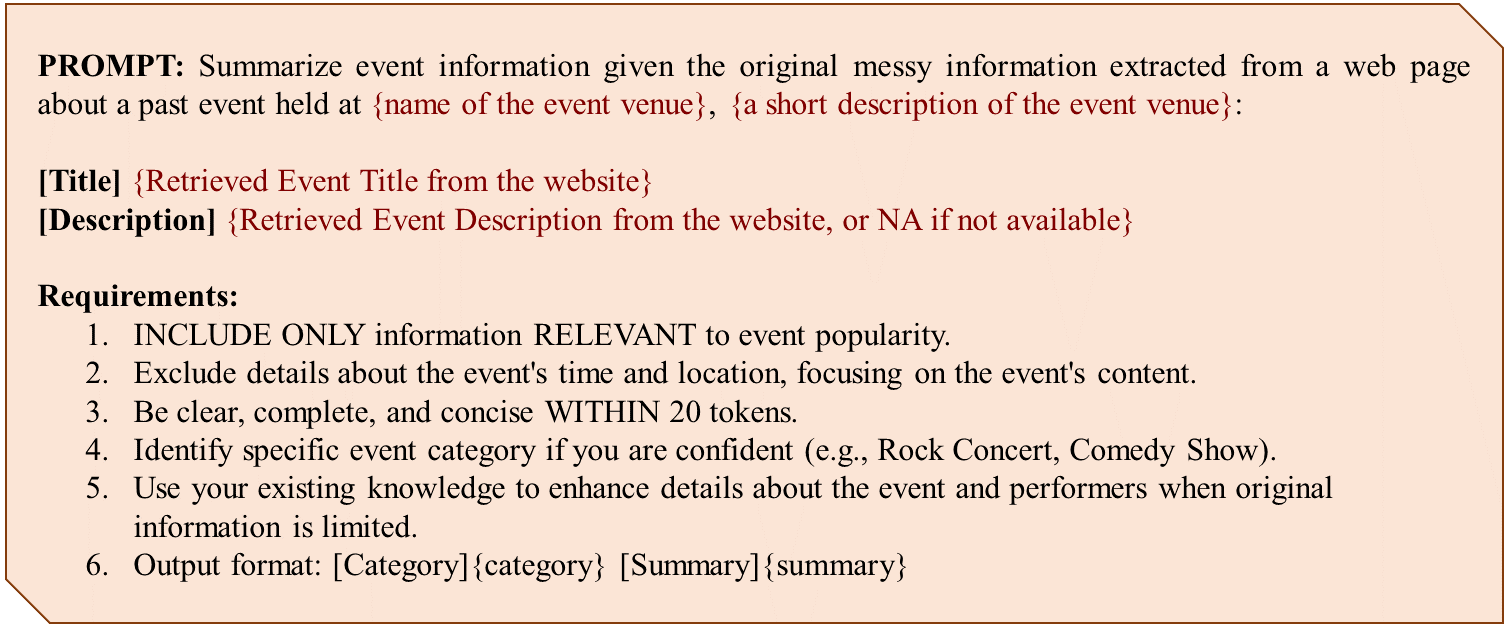}
  \caption{The prompt template used for formatting event features. Adapt for specific events by modifying the red-colored placeholders} 
\label{fig:event_prompt}
\end{figure}

Therefore, we design a prompt for event feature formatting, depicted in Figure~\ref{fig:event_prompt}. By inputting the event title (and its description if available), LLMs output a more informative event summary. Table~\ref{table:event_summary} shows 3 examples. Example A demonstrates that LLMs can adeptly condense lengthy and disorganized event descriptions into concise summaries, omitting non-essential details. Moreover, based on Examples B and C, it is clear that LLM has knowledge of a variety of specific nouns from basketball teams to electronic music bands. Equipped with such knowledge, they can identify event categories missing from the original titles.

\begin{table}[ht!]
  \scriptsize
  \centering 
  \caption{Examples of formatted event features using GPT-4}
    \begin{tabular}{p{0.15\linewidth} p{0.80\linewidth}}
    \toprule
    \multicolumn{2}{c}{\textbf{Example A}} \\
    Title & Charlie Wilson\\
    Description & Charlie Wilson is bringing his nationwide In It to Win It Tour to Barclays Center on March 29 with special guests Fantasia and Johnny Gill. Every ticket purchased online will come with one physical copy of Charlie’s new, upcoming In It To Win It album. Customers will receive a redemption email from Ticketmaster shortly after their purchase with instructions on how to secure their CD copy of Charlie’s new album. Any album redemption submitted before the album release date will be received by the customer on the official album release date. “I am excited about being on the road with my friends Grammy-Award winner, Fantasia, and four-time Grammy-Award nominee, Johnny Gill,” stated Wilson. “As usual, folks should come prepared for a party. I am looking forward to performing my new single, “Chills,” together with some music from my new album and my classic hits.” \\
    GPT-4's summary &  [Category] Concert [Summary] Charlie Wilson's In It to Win It Tour features special guests Fantasia and Johnny Gill. Every ticket includes a copy of his new album.\\
    \hline
    \multicolumn{2}{c}{\textbf{Example B}} \\
    Title & Brooklyn Nets vs. Dallas Mavericks\\
    Description & NA\\
    GPT-4's summary & [Category] NBA Basketball Game [Summary] A popular match between the Brooklyn Nets and Dallas Mavericks.\\
    \hline
    \multicolumn{2}{c}{\textbf{Example C}} \\
    Title & Porter Robinson, Giraffage, Lemâitre \\
    Description & NA\\
    GPT-4's summary & [Category] Electronic Music Concert [Summary] Porter Robinson, Giraffage, and Lemâitre performed, attracting fans of electronic music.\\
    \bottomrule
    \end{tabular}
  \label{table:event_summary}%
\end{table}%


\subsection{Mobility Feature Decomposition}\label{method:demand}

To more accurately discern the impact of public events on human mobility trends, we devise a method that separates travel demand into regular patterns (largely resulting from daily commuting) and irregular deviations (largely resulting from public events) \citep{rodrigues2016bayesian,chen2019subway, rodrigues2019combining}. This process begins with calculating a baseline for regular mobility patterns without events, utilizing a historical average model that considers travel demand from comparable past periods. For predictions of daily travel demand, this involves examining past mobility patterns on corresponding weekdays with no events. The irregular deviations are then computed as the difference between the actual travel demand and the estimated non-event demand. These calculated regular and irregular mobility patterns will be utilized as inputs for LLMs in the subsequent mobility prediction process, as will be introduced in the following section.

\subsection{Human Mobility Prediction}

Even with the processed event and mobility features, modeling human mobility under public events remains a challenge for current statistical and machine learning models for two main reasons. First, public events affect human mobility in complex ways, involving various factors such as the event category, performer's popularity, and event's timing, many of which are expressed as unstructured text not easily incorporated into traditional models. Second, the infrequent and unique nature of events makes it difficult to learn their impact on human mobility with sparse historical data. However, the recent advancements in LLMs offer a potential solution. LLMs have demonstrated the ability to tackle complex, multi-step reasoning tasks, such as solving mathematical problems, an ability they likely acquired through training on programming codes \citep{zhao2023survey}. Moreover, LLMs are effective at learning new tasks with very few examples \citep{brown2020language}, indicating their potential capability to infer event-induced travel demand by learning from limited historical events.

\begin{figure}[ht!]
  \centering
  \includegraphics[width=\linewidth]{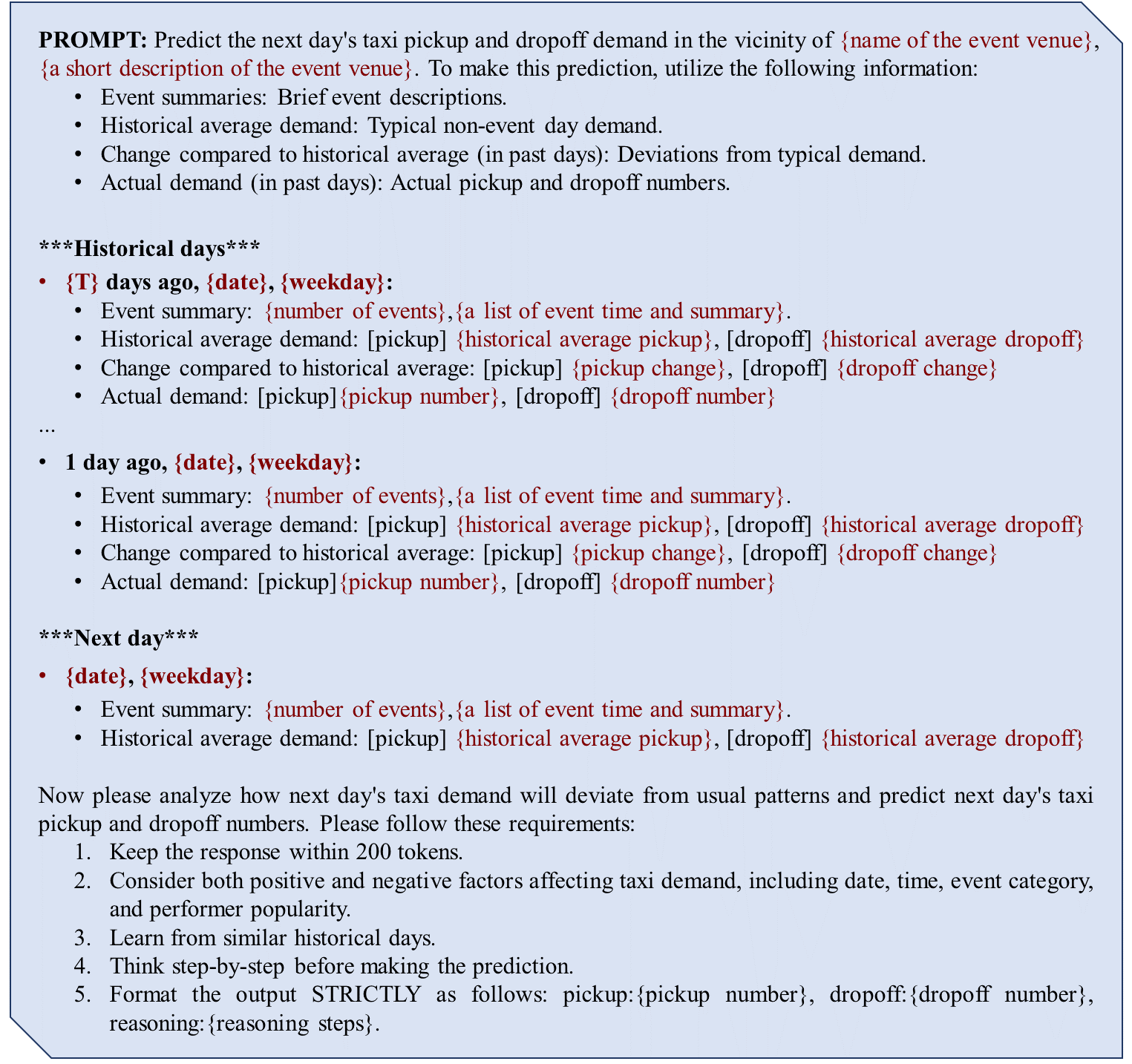}
  \caption{The prompt template used for travel demand prediction. Adapt for specific event venues and dates by modifying the red-colored placeholders.} 
\label{fig:predict_prompt}
\end{figure}

In this study, we utilize LLMs as human mobility predictors, using processed event and mobility features as introduced in Sections~\ref{method:event} and \ref{method:demand} as input. To harness the reasoning and few-shot learning capabilities of LLMs, we carefully design the prompt, as shown in Figure~\ref{fig:predict_prompt}. The prompt starts with an instruction detailing the task's input and desired output. This is followed by a presentation of historical event and mobility characteristics, coupled with the next time step's event specifics and estimated regular mobility patterns if no event occurs. The prompt concludes with a set of guidelines on how to approach the prediction task, like ``considering both positive and negative factors affecting travel demand, including date, time, event category, and performer popularity" and ``learning from similar historical days''.

Moreover, we ask LLMs to not only output predicted inflow and outflow values, but also the logical progression of their thought process leading to these predictions. This is achieved by instructing the model to ``think step-by-step before making the prediction'', as suggested by the chain-of-thought (CoT) prompting strategy \citep{wei2022chain}. Utilizing the CoT technique serves a dual purpose: first, it has been empirically validated to enhance performance across diverse tasks \citep{wei2022chain}; second, it facilitates the generation of predictions that are interpretable, as it articulates the reasoning in a format that is understandable and coherent to humans.

\section{Case Study} \label{sec:experiments}

\subsection{Data Description} \label{exp:data}

To verify the effectiveness of our proposed model, we conducted a case study on Barclays Center in New York City (NYC) using multi-source data for both public events and human mobility. Specifically, Barclays Center is chosen because of its role as a large-size venue for diverse public events and the availability of its complete historical event records online. Located in Downtown Brooklyn, Barclays Center is a multipurpose indoor arena that holds a variety of events such as concerts, sports games, and political gatherings. It is also the home ground for two American professional basketball teams, Brooklyn Nets and New York Liberty. With its capacity to seat approximately 19,000 spectators for basketball and up to 16,000 for concerts, the venue frequently attracts large crowds, thereby significantly impacting local mobility patterns. Our analysis utilizes taxi trip data to assess human mobility flows, owing to the point-to-point nature of taxi services, which enables us to accurately measure the flow of passengers to and from the venue. More details on the datasets we used are provided below.

\begin{figure}[ht!]
  \centering
  \includegraphics[width=\linewidth]{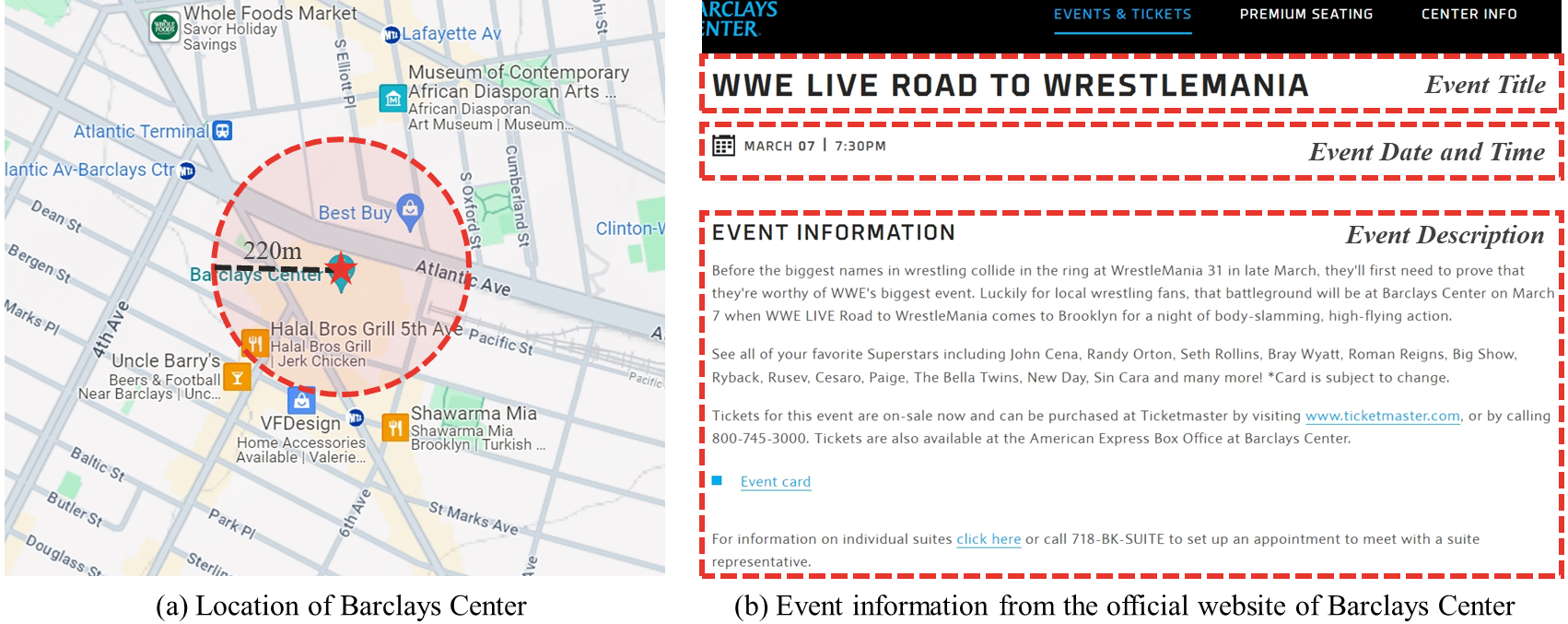}
  \caption{Data Description of Barclays Center} 
\label{fig:data_BC}
\end{figure}

\begin{figure}[ht!]
  \centering
  \includegraphics[width=\linewidth]{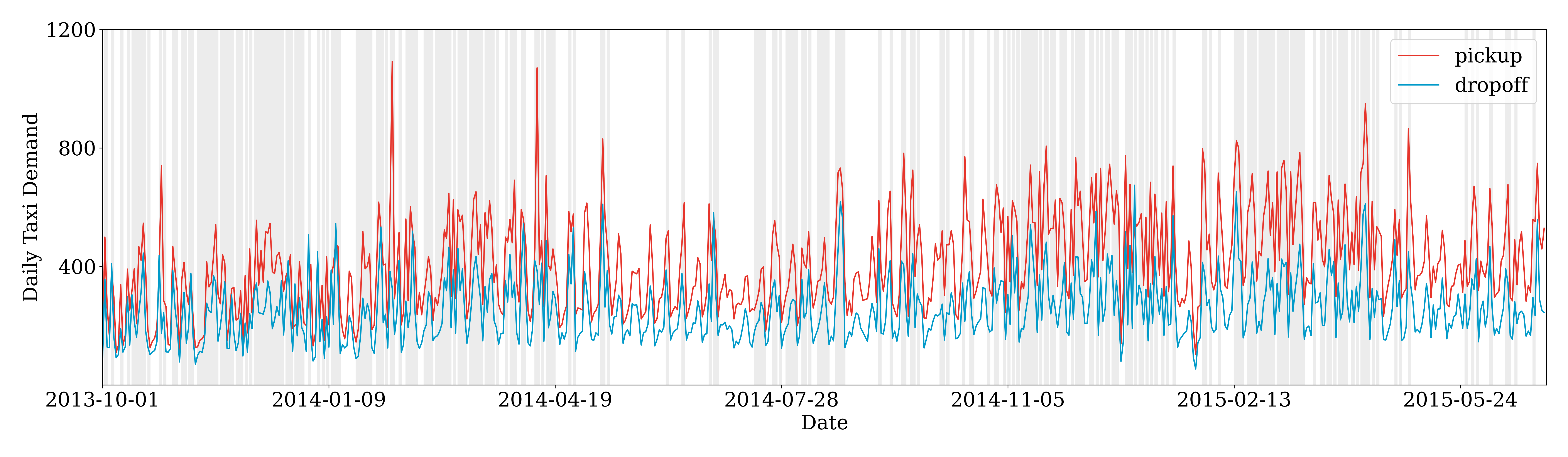}
  \caption{Temporal dynamics of daily taxi ridership in the vicinity of Barclays Center, with event days denoted by a grey background and non-event days by a white background.} 
\label{fig:temporal_demand}
\end{figure}

\textit{NYC Taxi Data}: The taxi trip data was collected from NYC Open Data\footnote{{\href{https://data.cityofnewyork.us/Transportation/Taxi/mch6-rqy4}{https://data.cityofnewyork.us/Transportation/Taxi/mch6-rqy4}}}, 
which provides detailed records of yellow and green taxi trips in New York City. We use data from 2013-10-01 to 2015-06-30 for experiments, due to the availability of exact pickup and dropoff geographic coordinates and the associated timestamps within this period. For our analysis, we identified taxi trips that began or ended within a 220m radius around Barclays Center, a distance chosen to encompass all adjacent roads (see Figure~\ref{fig:data_BC}(a)). These trips were subsequently aggregated into daily intervals, as illustrated in Figure~\ref{fig:temporal_demand}. There is a predominant trend of higher taxi pickups than dropoffs in proximity to Barclays Center, suggesting a preference among attendees to opt for alternative modes of transportation for arrival while favoring taxis for departure. Notably, event days at Barclays Center demonstrate a marked increase in taxi usage compared to non-event days, indicative of the larger crowd presence at these events. Furthermore, the magnitude of taxi ridership fluctuates significantly across different event days, presumably influenced by the nature of the events and the appeal of the performers. Therefore, an in-depth analysis of event-specific characteristics is imperative for accurately modeling their impact on taxi ridership.

\begin{figure}[ht!]
  \centering
  \includegraphics[width=0.5\linewidth]{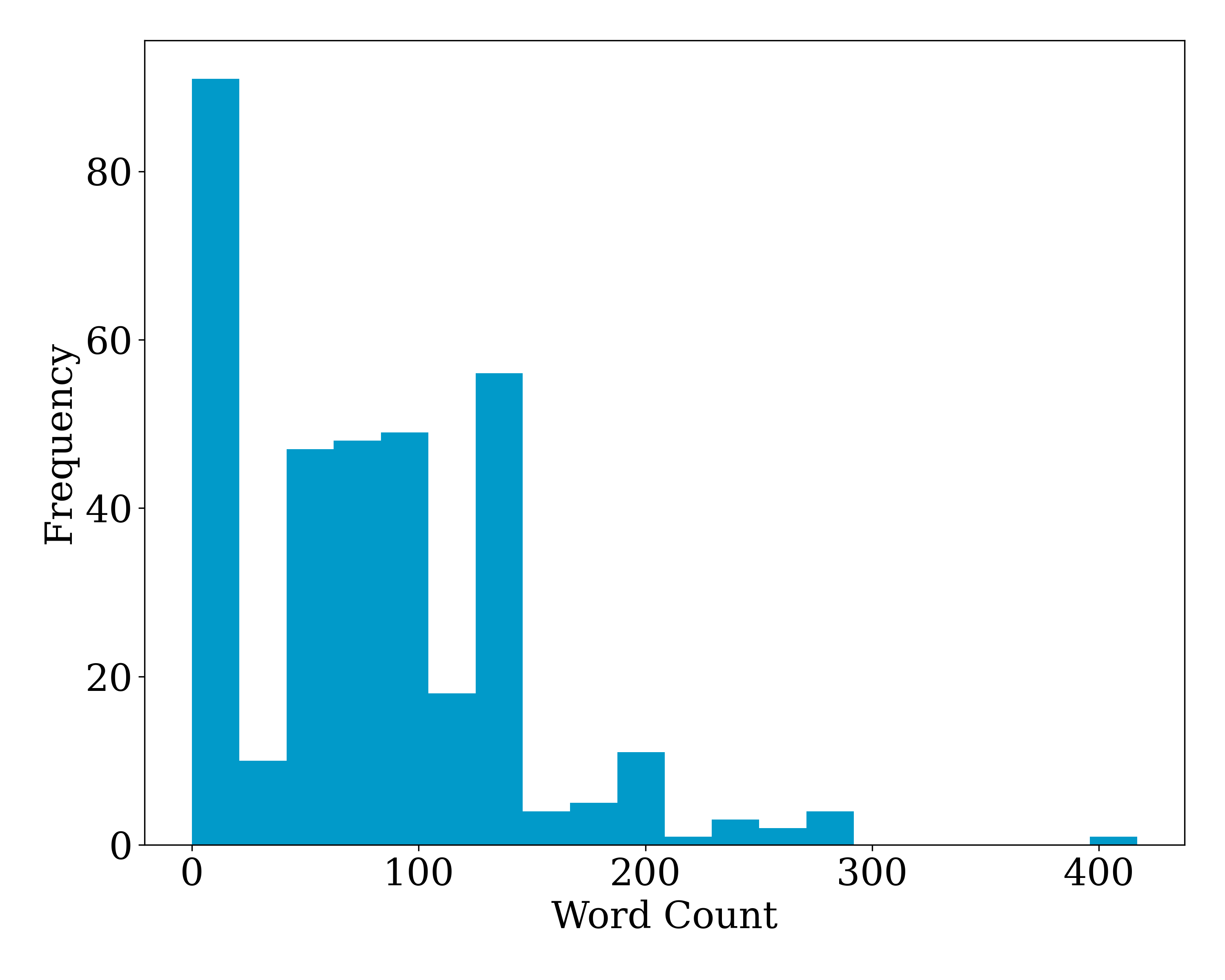}
  \caption{Distribution of word counts for event description data} 
\label{fig:word_cnt}
\end{figure}

\textit{Event Data}: The event data was scraped automatically from the official Barclays Center website\footnote{{\href{https://www.barclayscenter.com/events-tickets/event-calendar}{https://www.barclayscenter.com/events-tickets/event-calendar}}}, which maintains comprehensive records of past events. The website lists each event's title, scheduled start and end times, and occasionally more elaborate descriptions (see Figure~\ref{fig:data_BC}(b)). During our study period, we gathered data on 350 events spanning 285 days. To examine the variation in the extent and detail of event descriptions, we analyzed the word count distribution, as depicted in Figure~\ref{fig:word_cnt}. It can be clearly seen that while a notable portion of events lacks descriptions, others are associated with detailed narratives exceeding 400 words in some cases. It is thus important to format the event data to a consistent format and length to facilitate further human mobility analysis and prediction. Note that the majority of event venues in New York City list only upcoming events on their official websites. This limitation is a key reason behind our decision to focus exclusively on Barclays Center for our experiments.

\subsection{Experiment Setups} \label{exp:baseline}

In our experiments, we utilize GPT-4\footnote{{\href{https://openai.com/gpt-4}{https://openai.com/gpt-4}}}, one of the most advanced LLMs developed by OpenAI. To ensure result consistency across tests, we set the temperature parameter of GPT-4 to zero. The performance of LLM-MPE is compared with several representative statistical and machine learning models for time series prediction to assess its efficacy:
\begin{itemize}[noitemsep]
    \item \textbf{Linear Regression (LR)}: a regression model that assumes linear relationships between input and target variables. 
    \item \textbf{Gradient-Boosted Decision Trees (GBDT)}: an ensemble machine learning algorithm based on decision trees that can capture nonlinear relationships between input and target variables.
    \item \textbf{Feed Forward Network (FNN)}: a deep learning architecture consisting of an input layer, an output layer, and several hidden layers in between.
    \item \textbf{Recurrent Neural Network (RNN)}: a type of deep neural networks adept at capturing sequential dependencies in time series data. 
\end{itemize}

For all models in our study, we set the historical time window $T$ to 28 (days). Unlike LLMs, statistical and machine learning models require a dedicated training dataset for optimizing their parameters. Therefore, we divide our dataset into a training set from 2013-10-01 to 2014-06-30, and a test set from 2014-07-01 to 2015-06-30. LLM-MPE is only deployed on the test set to facilitate model comparison. 

For model inputs, we incorporate both historical mobility and event data for all baseline models to ensure a fair comparison. For event features, we include event count, time, and textual descriptions similar to LLM-MPE. However, while these features are easily integrated into LLMs using a natural language format, it is challenging to directly incorporate unstructured data like the event time and description in statistical and machine learning models. To address this issue, we represent event time as an occurrence distribution vector over different times in a day and textual descriptions as a topic distribution vector using Latent Dirichlet Allocation (LDA) following the practice of \cite{rodrigues2016bayesian}. For mobility features, we decompose human mobility into regular and irregular patterns using the same method in LLM-MPE. The baseline models then focus on predicting future irregular deviations instead of overall demand values. Our experiments show that the incorporated event features and irregular deviations can improve the prediction performance of both LLM-MPE and baseline models. 

The model performance is evaluated using the following metrics:
\begin{equation}
RMSE = \sqrt{\frac{1}{N}\sum_{n=1}^{N}{(Y_n - \hat{Y}_n)^2}}, \end{equation}
\begin{equation}
MAE = \frac{1}{N}\sum_{n=1}^{N}{|Y_n - \hat{Y}_n|},
\end{equation}
\begin{equation}
MAPE = \frac{100\%}{N}\sum_{n=1}^{N}{\frac{|Y_n - \hat{Y}_n|}{Y_n}},
\end{equation}
\begin{equation}
R^2 = 1 - \frac{\sum_{n=1}^{N}{(Y_n - \hat{Y}_n)^2}}{\sum_{n=1}^{N}(Y_n - \bar{Y})^2}, 
\end{equation}
where $N$ denotes the number of testing instances, $Y_n$ and $\hat{Y}_n$ correspond
to true and predict demand values for the $n$-th instance, and $\bar{Y}$ denotes
the average value of $Y$. To clearly demonstrate how models perform under different conditions, we evaluate model performance on days with and without events separately. In our test set, there are 149 days associated with 181 events and 216 days that are event-free. For deep learning models, FNN and RNN, we execute each model 10 times and report the average performance.

\section{Results}

\subsection{Performance Comparison with Baseline Models}

Table~\ref{table:performance} summarizes the human mobility prediction performance of different models on the test set. For all models, the RMSE and MAE are significantly higher on days with public events compared to those without, likely due to the typically increased travel demand on event days. Conversely, days without events show higher $R^2$ values, indicating a more consistent and predictable mobility pattern. Compared with baseline models, LLM-MPE achieves the best performance on all evaluation metrics in all circumstances, especially on event days. This highlights the effectiveness of LLMs in processing the textual details of event descriptions and in capturing the complex influence of public events on travel demand. Additionally, even on days without events, LLM-MPE can notably outperform existing models. This may be attributed to the ability of LLM-MPE to utilize historical mobility data more effectively, aided by the information from event descriptions in past days.  

Among baseline models, GBDT performs better than LR, highlighting the limitations of traditional statistical models in identifying nonlinear relationships. It is interesting to note that FNN and RNN do not perform as well as GBDT in our case. This can be attributed to the limited amount of training data available. Deep learning models typically require a substantial volume of training data (e.g. over 10,000 data points) to yield accurate predictions. However, such large datasets are often not feasible when dealing with public events considering their infrequent and diverse nature. In our case, although our study period spans nearly two years, there are only 350 public events recorded in the dataset. This quantity is insufficient for deep learning models to learn meaningful correlations between public events and travel demand. Compared with the best baseline model GBDT, LLM-MPE can greatly improve the RMSE and MAE by 13.9\% and 17.8\% on event days, and by 4.1\% and 13.7\% on non-event days.

\begin{table}[ht!]
  \centering \footnotesize
  \caption{Performance comparison of LLM-MPE with baseline models}
    \resizebox{\linewidth}{!}{%
    \begin{tabular}{ccccccccccccc}
    \toprule
    \multirow{2}{*}{Models} & \multicolumn{4}{c}{All Days} & \multicolumn{4}{c}{Event Days} & \multicolumn{4}{c}{Non-Event Days} \\
    & RMSE & MAE & MAPE & $R^2$ & RMSE & MAE & MAPE & $R^2$& RMSE & MAE & MAPE & $R^2$\\
    \midrule
    LR & 91.692 & 68.700 & 0.217 & 0.702 & 115.377 & 89.144 & 0.210 & 0.550 & 70.883 & 54.596 & 0.221 & 0.728\\
    GBDT & 79.716 & 57.412 & 0.173 & 0.765 & 103.832 & 80.679 & 0.180 & 0.621 & 57.456 & 41.362 & 0.167 & 0.785\\
    FNN & 93.395 & 74.959 & 0.244 & 0.695 & 109.019 & 86.339 & 0.191 &  0.617 & 80.627 & 67.109 & 0.280 & 0.790\\
    RNN & 103.293 & 85.926 & 0.282 & 0.634 & 112.912 & 91.173 & 0.205 & 0.580 & 96.098 & 82.307 & 0.334 & 0.790\\ 
    LLM-MPE & \underline{\textit{71.128}} & \underline{\textit{48.171}} & \underline{\textit{0.136}} & \underline{\textit{0.816}} & \underline{\textit{89.374}} & \underline{\textit{66.279}} & \underline{\textit{0.149}} & \underline{\textit{0.715}} & \underline{\textit{55.127}} & \underline{\textit{35.681}} & \underline{\textit{0.127}} & \underline{\textit{0.799}}\\
    \bottomrule
    \end{tabular}}
  \label{table:performance}%
\end{table}%

\begin{figure}[ht!]
  \centering
  \includegraphics[width=\linewidth]{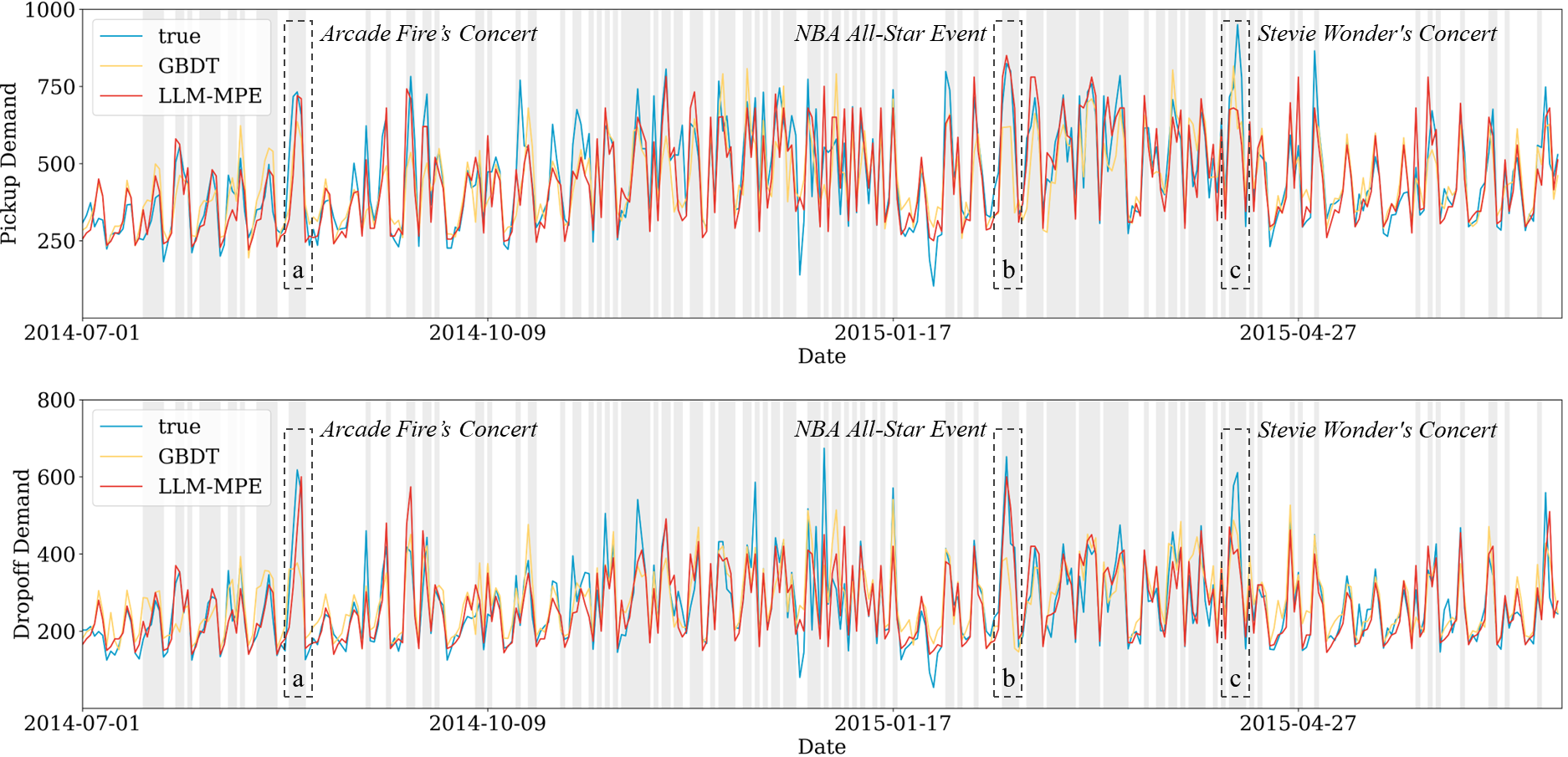}
  \caption{The true and predicted demand with LLM-MPE and XGBoost} 
\label{fig:predict_plot}
\end{figure}

To intuitively illustrate the human mobility prediction results, we visualize the actual travel demand values from our test set and the predictions made by LLM-MPE and GBDT. As shown in Figure~\ref{fig:predict_plot}, the prediction lines from LLM-MPE (colored red) closely follow the real human mobility trends (colored blue) in most instances. In contrast to the predictions by XGboost (shown in yellow), LLM-MPE more effectively mirrors the travel demand variations on days with events, particularly during major public events (as seen in Examples a and b). However, there are sporadic instances where our model either overpredicts or underpredicts the travel demand at peak times (as noted in Example c). This is understandable, given the multitude of interconnected factors influencing travel demand, which makes it a formidable task to precisely forecast human mobility for every event.

\subsection{Performance Comparison with Different Input Features}

To quantify the contribution of different features, we analyze how LLM-MPE performs with different sets of input features. In addition, considering the unique working mechanism of LLMs compared to other machine learning models, we use GBDT as an example to explore how these features affect such models differently. In both LLM-MPE and baseline models, we focus on three main event-related features: the count of events ($c$), event timings ($t$), and textual descriptions ($h$). In the case of LLM-MPE, a preprocessing step is applied to transform the raw, unstructured textual descriptions $h$ into a more standardized and concise format, referred to as $h'$. To verify their effectiveness, we perform several simplified versions of event features as listed below:
\begin{itemize}[noitemsep]
    \item No event features (NA): This version excludes all event features, relying solely on historical mobility data.
    \item Event count ($c$):  This version includes only the number of events as the event feature.
    \item Event count and timing ($c + t$): In this variant, textual descriptions are omitted. For LLM-MPE, this means removing event descriptions from the prompt. For GBDT, it involves excluding the topic distribution vector obtained from LDA.
    \item Event count, timing, and raw descriptions ($c+t+h$): Here, the original event descriptions are used as inputs for LLM-MPE, instead of the formatted versions.
\end{itemize}

In terms of mobility features, both LLM-MPE and the baseline models segment the original mobility data into ``regular'' daily travel patterns ($r$) and ``irregular'' event-induced deviations ($i$). To test the effectiveness of this approach, we create a modified version of LLM-MPE and GBDT:
\begin{itemize}[noitemsep]
    \item Original travel demand ($o$): In this variant, LLM-MPE excludes the historical average estimations and travel demand deviation data from the prompt. GBDT predicts future demand values directly, rather than focusing on the deviations.
\end{itemize}

Table~\ref{table:ablation} presents the outcomes of using various input features in LLM-MPE and GBDT. The experiments are based on days with events which are the main focus of this research. It can be found that all input features positively impact the performance of both models. Among event-related features, knowledge of the next day's event count notably reduces prediction errors, enhancing RMSE and MAE by 27.7\% and 30.3\% for LLM-MPE, and by 16.2\% and 17.6\% for GBDT, respectively. This again validates that the occurrence of events can significantly affect travel demand. Additionally, the inclusion of event timings modestly boosts the performance of both models across most evaluation metrics. When it comes to textual descriptions, incorporating the raw event descriptions can already improve the performance of LLM-MPE by a large margin, with a reduction of 22.0\% and 24.9\% in RMSE and MAE. For GBDT, including a topic distribution vector derived from LDA also enhances performance, albeit not as significantly as with LLM-MPE. This indicates that LLMs can better utilize event descriptions to enhance prediction accuracy due to their advanced understanding of natural language. The formatted event descriptions can further enhance LLM-MPE's performance, which can be explained that with explicitly identified event categories and consistent description formats, it is easier for LLMs to compare and distinguish event features across different days. As for mobility features, relying solely on travel demand values increases the prediction error. For LLM-MPE, this results in an 11.1\% and 16.1\% increase in RMSE and MAE, and for GBDT, the increase is 12.8\% and 11.0\%. This indicates the value of mobility feature decomposition, which helps models focus more on irregular deviations from regular mobility patterns.

\begin{table}[ht!]
  \centering \footnotesize
  \caption{Performance comparison of LLM-MPE and GBDT on event days with different input features}
    \resizebox{\linewidth}{!}{%
    \begin{tabular}{ccccccccccc}
    \toprule
    \multirow{2}{*}{Types} & \multirow{2}{*}{Fixed Setting} & \multirow{2}{*}{Features} & \multicolumn{4}{c}{LLM-MPE} & \multicolumn{4}{c}{GBDT}\\
    & & & RMSE & MAE & MAPE & $R^2$& RMSE & MAE & MAPE & $R^2$\\
    \midrule
    \multirow{4}{*}{Event} & \multirow{4}{*}{\makecell{Demand:\\$r+i$}} & NA & 167.321 & 134.034 & 0.277 & 0.455 & 130.397 & 101.882 & 0.213 & 0.540\\
    & & $c$ & 120.965 & 93.416 & 0.220 & 0.511 & 109.294 & 83.961 & 0.197 & 0.557\\
    & & $c+t$ & 114.793 & 87.232 & 0.212 & 0.552 & 107.081 & 84.272 & 0.197 & 0.575\\
    & & $c+t+h$ & 91.847 & 69.889 & 0.158 & 0.700 & \underline{\textit{103.389}} & \underline{\textit{80.205}} & \underline{\textit{0.179}} & \underline{\textit{0.624}} \\
    & & $c+t+h'$ & \underline{\textit{89.374}} & \underline{\textit{66.279}} & \underline{\textit{0.149}} & \underline{\textit{0.715}} & - & - & - & - \\
    \hline
    \multirow{4}{*}{Demand} & \multirow{2}{*}{\makecell{Event:\\$c+t+h'$}} & $o$ & 99.215 & 76.960 & 0.183 & 0.659 & - & - & - & -\\
    & & $r+i$ & \underline{\textit{89.374}} & \underline{\textit{66.279}} & \underline{\textit{0.149}} & \underline{\textit{0.715}} & - & - & - & - \\
    & \multirow{2}{*}{\makecell{Event:\\$c+t+h$}} & $o$ & - & - & - & -& 116.641 & 89.016 & 0.186 & 0.578\\
    & & $r+i$ & - & - & - & - & \underline{\textit{103.389}} & \underline{\textit{80.205}} & \underline{\textit{0.179}} & \underline{\textit{0.624}} \\
    \bottomrule
    \end{tabular}}
  \label{table:ablation}%
\end{table}%





\subsection{Interpretability of LLMs}

In addition to human mobility prediction, LLMs can also provide step-by-step reasonings regarding why they make such predictions. To demonstrate the interpretability of LLMs, we examine several examples as detailed in Table~\ref{table:interpretation}. Each example includes the date to be predicted, upcoming events, the actual and forecasted travel demand values by GBDT and LLM-MPE, and the reasoning process as articulated by LLM-MPE. These examples show that LLM-MPE can offer clear and detailed explanations for its predictions. By looking into the generated reasoning steps, we observe that LLM-MPE is adept at capturing complex temporal patterns, including not only weekly regularity but also daily trends. For instance, in Example I, LLM-MPE observed that ``the travel demand on previous days with similar events have demonstrated a slight decrease in pickups and a slight increase in dropoffs", which was then leveraged in its prediction. When there are prior days with identical events, LLM-MPE can identify these days and use the past travel demand changes caused by the same events to refine its predictions, as seen in Example II. Conversely, for new events that are unseen before, such as the NBA All-Star Event in Example III, LLM-MPE is still capable of finding comparable historical events (like regular NBA games). It then assesses the similarities and differences between these events to adjust its predictions. For instance, in Example III, LLM-MPE points out that ``All-Star Event is likely to be even more popular than a regular game, and thus an even larger increase in travel demand is expected". 

\begin{table}[ht!]
  \scriptsize
  \centering 
  \caption{Examples of prediction and reasoning results by LLM-MPE}
    \begin{tabular}{p{0.16\linewidth} p{0.79\linewidth}}
    \toprule
    \multicolumn{2}{c}{\textbf{Example I}} \\
    Date & 2014-07-20, Sunday\\
    Event Summary & 2 events: [time] 13:00-16:00,17:00-20:00 [Category] Dinosaur Show [Summary] The spectacular production, Walking With Dinosaurs features 20 life-size dinosaurs, including the Tyrannosaurus rex, Plateosaurus, and Brachiosaurus\\
    True Demand & [pickup] 399 [dropoff] 253 \\
    GBDT’s results & [pickup] 483 [dropoff] 304 \\
    LLM-MPE’s results & [pickup] 410 [dropoff] 260 [reasoning] The historical average demand for a Sunday is 426 pickups and 247 dropoffs. \textit{The event, "Walking With Dinosaurs", has been running for the past few days. Comparing the demand on these days with the historical average, we see a slight decrease in pickups and a slight increase in dropoffs.} Considering the event is still popular but has been running for a few days, we can expect a similar pattern. Therefore, I predict a slight decrease in pickups to 410 and a slight increase in dropoffs to 260. \\
    \hline
    \multicolumn{2}{c}{\textbf{Example II}} \\
    Date & 2014-07-25, Friday\\
    Event Summary & 1 event: [time] 19:30-22:30 [Category] Pop Music Concert [Summary] International superstar Katy Perry's PRISMATIC WORLD TOUR supports her PRISM album, which debuted at number one on The Billboard 200. The album includes her biggest single "Roar" and 13th Top Ten hit "Dark Horse". Special guest is Capital Cities.\\
    True Demand & [pickup] 555 [dropoff] 354 \\
    GBDT’s results & [pickup] 462 [dropoff] 198 \\
    LLM-MPE’s results & [pickup] 562 [dropoff] 353 [reasoning] The event is a popular pop music concert by Katy Perry, which is likely to attract a large crowd. Looking at the historical data, on 2014-07-24, a similar event by the same artist led to a significant increase in both pickup and dropoff numbers. The pickup demand increased by 231 and the dropoff demand increased by 146 compared to the historical average. \textit{Considering that the event on the next day is similar, we can expect a similar increase in demand. Therefore, the predicted pickup demand is 331 (historical average) + 231 (increase due to event) = 562. The predicted dropoff demand is 207 (historical average) + 146 (increase due to event) = 353.}\\
    \hline
    \multicolumn{2}{c}{\textbf{Example III}} \\
    Date & 2015-02-14, Saturday\\
    Event Summary & 1 event: [time] 20:30-23:30 [Category] NBA All-Star Event [Summary] State Farm All-Star Saturday Night features NBA's top stars, hosted by Anthony Anderson, with performances by Ella Henderson.\\
    True Demand & [pickup] 824 [dropoff] 652 \\
    GBDT’s results & [pickup] 618 [dropoff] 390 \\
    LLM-MPE’s results & [pickup] 850 [dropoff] 600 [reasoning] First, we note that the event is an NBA All-Star Event, which is likely to draw a large crowd due to the popularity of the NBA and the star power involved. \textit{Looking at similar days, we see that NBA games consistently increase taxi demand significantly.} For example, 15 days ago, an NBA game increased pickup demand by 410 and dropoff demand by 206. 12 days ago, another NBA game increased pickup demand by 238 and dropoff demand by 141. Yesterday, an NBA game increased pickup demand by 307 and dropoff demand by 249. \textit{Given that the All-Star Event is likely to be even more popular than a regular game, we can expect an even larger increase in demand.} Therefore, we predict that the pickup demand will increase by around \underline{\textit{296 (average of 410, 238, and 307)}} from the historical average of 554, and the dropoff demand will increase by around \underline{\textit{293 (average of 206, 141, and 249)}} from the historical average of 307. \\
    \bottomrule
    \end{tabular}
  \label{table:interpretation}%
\end{table}%

While LLM-MPE has demonstrated impressive reasoning abilities in previous examples, it is important to acknowledge that it can occasionally generate inaccurate information. For instance, as underlined in Example III, LLM-MPE makes a wrong mathematical calculation: the average of 410, 239, and 307 should be 319, not 296; and the correct average of 206, 141, and 249 is 199, not 293. This suggests the possibility of LLMs generating inaccurate and misleading information. As a result, users need to be cautious of such potential errors when utilizing these models.

\section{Discussion: Limitations of LLMs}

In the previous sections, we demonstrated the superior ability of LLMs regarding formatting unstructured textual data, making demand predictions, and generating detailed explanations. However, certain drawbacks currently hinder the widespread application of LLMs in large-scale human mobility applications. Therefore, this research should be considered as an initial exploration into the potential of integrating LLMs for human mobility predictions during public events, rather than a complete, real-world solution. This section aims to delve into the current limitations of LLMs, offering insights that could be instrumental for future research in enhancing LLM applications in human mobility prediction.

Firstly, as previously mentioned, even advanced LLMs like GPT-4 can sometimes produce incorrect information, a phenomenon known as ``hallucination'' in LLMs. This issue is particularly concerning from an ethical standpoint, as these inaccuracies can lead to irresponsible or even harmful recommendations, potentially impacting human safety and well-being \citep{bang2023multitask}. Secondly, LLMs' vast knowledge base, essential for interpreting event descriptions and making predictions, may not always be up-to-date due to training data limitations. This shortcoming can impede their understanding of the most recent public events. For example, as of November 2023, GPT-4 was only informed up to April 2023, thereby lacking knowledge of new trends or emerging artists. Thirdly, most state-of-the-art LLMs, including GPT-4, are not open source and incur costs for usage. This could lead to significant expenses when processing large volumes of data. A fourth challenge involves prediction inefficiency due to the necessity of invoking the OpenAI API for every single prediction instance. In our case, GBDT takes approximately 2.4 seconds to generate predictions for all test samples, while LLM-MPE takes 12.7 seconds to generate predictions for a single test sample. This can be particularly problematic in scenarios involving large-scale data, such as city-wide analyses, or when high-frequency predictions are needed, such as forecasting travel demand every 5 minutes. Lastly, our study, which concentrates on a specific event venue, does not consider the spatial interactions between various venues. In citywide prediction scenarios, these interactions can be crucial. Presently, there is no widely accepted method to encode spatial relationships in a manner that LLMs can understand. Furthermore, even if spatial relationships are articulated in natural language, it's unclear how effectively LLMs can comprehend and rationalize based on this information.

To mitigate these challenges, several solutions can be explored in future works. First, there is a need for more stringent AI regulations and guidelines to ensure responsible use. In applications such as human mobility prediction, transportation system operators should be fully aware of the potential risks, such as inaccuracies and outdated information, and establish procedures to review and refine their outputs before making decisions. Second, we encourage the development of open-source LLMs, which would not only be cost-effective but also give users full control over their models, enabling adaptations to specific use cases. Third, tackling the inefficiency in LLM-based predictions could involve combining LLMs with traditional machine-learning models. For example, in our case study, as conventional machine learning models show comparable effectiveness on days without events, it might be practical to employ these models for non-event days and reserve LLMs for days with events. Last, for incorporating spatial dependencies in LLMs, two strategies could be considered. One approach is to articulate spatial relationships in natural language (e.g., describing one venue's proximity to another). Alternatively, LLMs could be combined with domain-specific deep neural networks, like graph neural networks (GNNs). This could involve using LLMs to convert textual data into a latent space, which could then be used as node attributes in GNNs \citep{chen2023exploring}.
 
\section{Conclusion} \label{sec:conclusion}
 
This study introduces the use of Large Language Models (LLMs) to enhance human mobility prediction under public events. LLMs, compared to traditional statistical and machine learning techniques, excel in processing textual data, learning from minimal data, and generating human-readable explanations. Specifically, a prediction framework called LLM-MPE is presented, consisting of two primary phases: feature development and mobility prediction. In the feature development phase, unstructured textual data obtained from the internet is transformed into a concise, uniform format. Additionally, historical mobility data is split into regular patterns (largely attributed to daily commuting) and irregular deviations (largely attributed to public events). During the mobility prediction phase, a crafted prompt template is carefully designed to guide LLMs in making predictions based on past mobility and event characteristics, and in articulating the reasoning behind these predictions. Our case study, based on New York City taxi trip data and online event information from Barclays Center, shows that LLM-MPE surpasses existing machine learning models, particularly on days with events. Ablation studies reveal the critical role of the textual data from event descriptions in LLM-MPE's performance. Moreover, we analyze the explanatory steps provided by LLMs, highlighting their ability to offer clear and detailed justifications for their predictions. However, despite the promising use of LLMs in improving predictions of human mobility during public events, there is still considerable progress to be made before achieving a fully viable solution for real-world applications. Currently, LLMs face challenges such as the production of inaccurate information, reliance on outdated data, high operational costs, efficiency constraints, and a lack of capability to understand spatial relationships. These challenges restrict their immediate use in large-scale human mobility tasks.

This research can be further improved in the following aspects. First, while our experiments are centered around a specific event venue, our proposed method is generalizable and can be easily adapted to other venues, provided that the human mobility data and event information are accessible. Future studies can test LLM-MPE on other venues with varying locations, sizes and event types, to further validate the consistency of the model performance. Second, our current focus is on forecasting the inflow and outflow of travel demand. Future research could broaden this scope to include various dimensions of human mobility during public events, such as analyzing origin-destination flows, specific time of travel, transportation mode choices, and other individual travel decisions. Third, in addition to public events, predicting travel demand during unforeseen incidents, such as traffic accidents, is also crucial. The unpredictability of the timing and location of these events makes this task more challenging. Nonetheless, past studies have shown the feasibility of detecting sudden traffic occurrences from website or social media data through textual data analysis \citep{xu2018sensing}. Therefore, employing LLMs to swiftly identify such unforeseen events is likely to help improve real-time predictions of human mobility flows in these scenarios.

\section*{Acknowledgements}
This research is supported by National Natural Science Foundation of China (NSFC 42201502).






\bibliographystyle{model5-names2}\biboptions{authoryear}
\bibliography{main}







\end{document}